\documentclass[10pt,twocolumn,letterpaper]{article}

\usepackage{cvpr}              

\usepackage{graphicx}
\usepackage{amsmath}
\usepackage{amssymb}
\usepackage{booktabs}
\usepackage[accsupp]{axessibility}

\usepackage[pagebackref,breaklinks,colorlinks]{hyperref}

\usepackage[capitalize]{cleveref}
\crefname{section}{Sec.}{Secs.}
\Crefname{section}{Section}{Sections}
\Crefname{table}{Table}{Tables}
\crefname{table}{Tab.}{Tabs.}

\def\cvprPaperID{63} 
\def\confName{CVPR}
\def\confYear{2022}

\begin{document}

\title{Alpha Matte Generation from Single Input for Portrait Matting}

\author{Dogucan Yaman\textsuperscript{1} \qquad Hazım Kemal Ekenel\textsuperscript{2} \qquad Alexander Waibel\textsuperscript{1,3}\\
\textsuperscript{1}Karlsruhe Institute of Technology, \textsuperscript{2}Istanbul Technical University, \textsuperscript{3}Carnegie Mellon University\\
{\tt\small \{dogucan.yaman, alexander.waibel\}@kit.edu, ekenel@itu.edu.tr}
}
\maketitle

\begin{abstract}

In the portrait matting, the goal is to predict an alpha matte that identifies the effect of each pixel on the foreground subject. Traditional approaches and most of the existing works utilized an additional input, e.g., trimap, background image, to predict alpha matte. However, (1) providing additional input is not always practical, and (2) models are too sensitive to these additional inputs. To address these points, in this paper, we introduce an \textit{additional input-free} approach to perform portrait matting. We divide the task into two subtasks, segmentation and alpha matte prediction. We first generate a coarse segmentation map from the input image and then predict the alpha matte by utilizing the image and segmentation map. Besides, we present a segmentation encoding block to downsample the coarse segmentation map and provide useful feature representation to the residual block, since using a single encoder causes the vanishing of the segmentation information. We tested our model on four different benchmark datasets. The proposed method outperformed the MODNet and MGMatting methods that also take a single input. Besides, we obtained comparable results with BGM-V2 and FBA methods that require additional input.


\end{abstract}

\section{Introduction}
\label{sec:intro}
Image matting has become a popular research topic in the computer vision research area. The main purpose is to distinguish background and foreground to obtain foreground objects as accurately as possible. Therefore, the task is to generate an alpha matte that contains alpha values, namely opacity values, between $[0, 1]$ for each pixel to represent the effect of the foreground over the final image. In addition to this, portrait matting, which is a subtopic of image matting, focuses on generating alpha matte to obtain the subject itself, instead of the generic objects, from an input image or a video frame. There are numerous application areas of portrait matting, such as image/video editing, changing background which is quite common in video conference applications, and video/movie post-production.


There are various challenges in the portrait matting problem due to the complex visual details of a person's body, e.g., the borders around the body, the hair, and the clothes, particularly if the hair flutters and the clothes have some opacity. The matting problem can be formulated as follows:

\begin{equation}
    I_i = \alpha_i F_i + (1-\alpha_i) B_i
    \label{equation_combination}
\end{equation}
where $i$ represents each pixel in an image I, alpha represents alpha value for the corresponding pixel in the alpha matte $ \alpha $, and $ F $ and $ B $ are foreground and the new background images, respectively. 

In the traditional approach, the alpha matte is generated using an image and a trimap which represents the foreground, background, and unknown areas on the image. 
The basic idea is to enhance the unknown areas in the trimap, which are generally problematic parts of the subject, e.g., the area around the subject's body, to get a more accurate alpha matte. 
The predefined foreground and background areas are not changed. On the contrary, several latest works \cite{liu2020boosting,ke2020green,sengupta2020background} propose not to use a trimap, since creating trimap is a time-consuming procedure and needs expert annotators. Instead, some works employ original image and coarse annotated segmentation mask to generate a fine-grained alpha matte \cite{sengupta2020background, xu2021virtual}. Moreover, recent work focuses on using an input image and its background to produce alpha matte without using any other information \cite{lin2020real}, while other works utilize only the input image to achieve fine-grained alpha matte \cite{liu2020boosting,ke2020green,yu2021cascade,mgm} and predict trimap to use in the alpha matte prediction \cite{sun2021deep}. 

One of the crucial challenges is posed by the distribution of the background and foreground of an image. It is an extremely severe case when the background distribution is considerably similar to the foreground distribution. Besides, if the background distribution has a large variance, it is another compelling case to handle the discrimination of background and foreground subject. Yet another challenge arises from the illumination conditions of the input image since the background matting models are sensitive to the illumination distribution. In particular, the alpha matte prediction models are prone to generate coarse, even worse, outputs under the cases of underexposure and overexposure.

In this work, we aim to enhance the quality of the generated alpha matte to extract the person, since fine-grained details of the subjects are the main challenges in the portrait matting task. To alleviate the problem, we handled it using two consecutive stages, which are person segmentation and alpha matte generation. We employed DeepLabv3+ \cite{deeplabv3_plus} for person segmentation and a generative adversarial network-based (GAN) alpha matte prediction model. 
While the first network takes an input image and produces the segmentation map, the alpha generation network employs the output of the segmentation network and foreground subject, which is obtained by multiplication of the input image and predicted segmentation map. In the end, the refinement block receives the predicted alpha matte to refine the details. Our contributions can be summarized as follows:

\begin{itemize}
    \item We propose a two-stage portrait matting network, that consists of a SOTA person segmentation network DeepLabv3+ and subsequently a conditional GAN-based alpha matte prediction module, without using an additional input as trimap, background image, etc.
    \item We present \textit{segmentation encoding block} to encode the predicted segmentation map and the foreground subject. The idea is to obtain the feature representation of the segmentation map and foreground subject independently of the input, and inject it into the residual block as well as decoder layers along with the depth. We observed that using an independent encoder, instead of encoding concatenation of all inputs with a single encoder, provides better feature representation.
    \item We propose \textit{border loss} to penalize the errors around the subject more, since it is more likely to have errors in the prediction due to difficulties, such as hair. We also present \textit{alpha coefficient loss} to evaluate only the pixels that have neither 0 nor 1 value in the alpha matte.
    
    
\end{itemize}



\section{Related Work}
Although person segmentation can be employed to extract the subject from an image as well as replace the background, it is not adequately accurate to eliminate the background and its effects on the subject. Therefore, alpha matte generation is a more accurate approach  for background replacement or portrait matting.  

\begin{figure*}
\begin{center}
\includegraphics[scale=0.15]{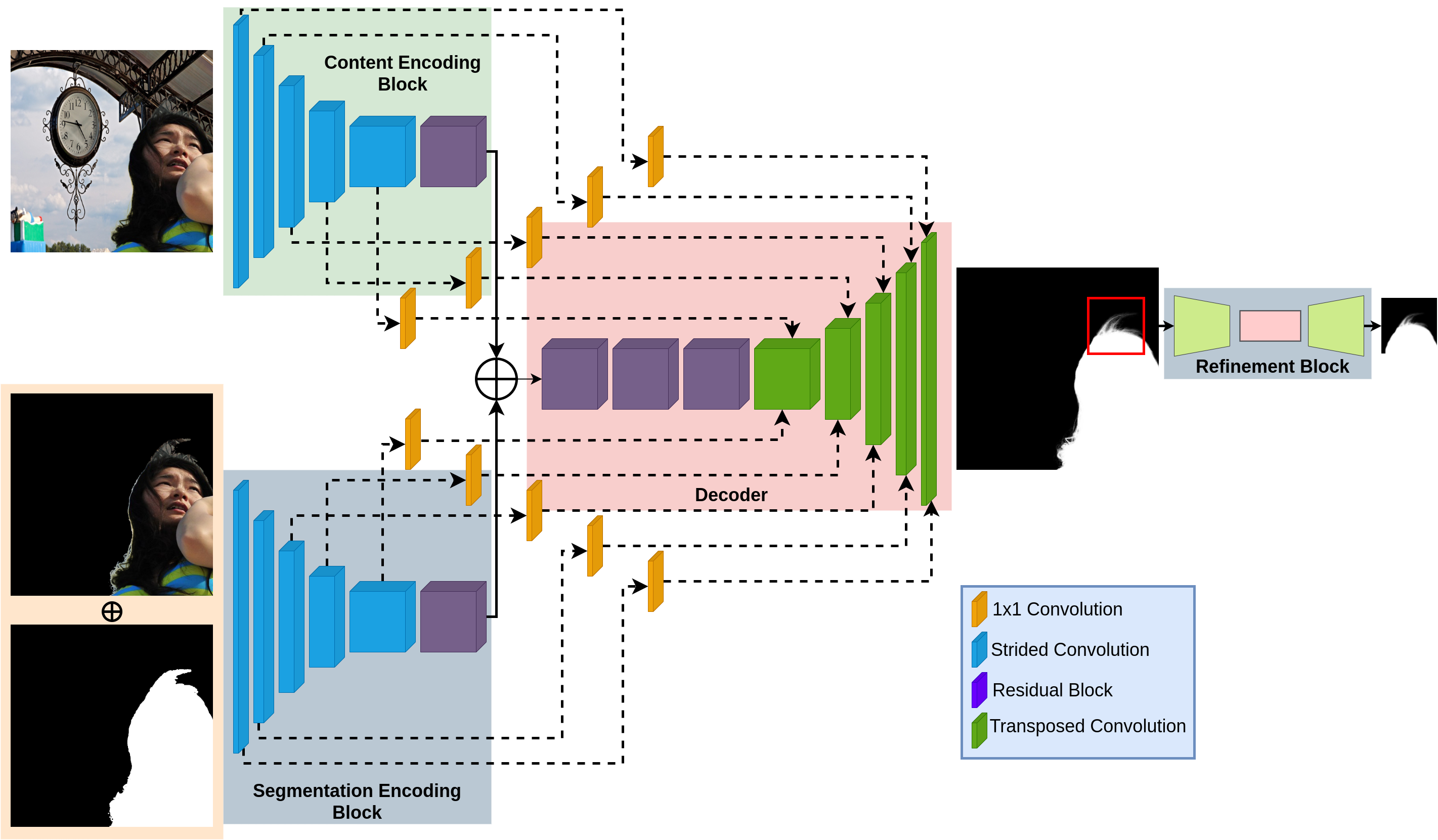}
\end{center}
   \caption{
   Proposed model. First of all, DeepLabv3+ \cite{deeplabv3_plus} model works and produces a segmentation map. After that, the visualized system starts to work using the input image and the predicted coarse segmentation map. While the content encoding block encodes the input image to provide feature maps for the decoder network, the segmentation encoding block employs the combination of the predicted segmentation map and foreground subject that is obtained by multiplication of the input image and the predicted segmentation map. Besides, residual connections between encoders' layers and decoder's layers are effective to preserve the information. After each encoders' layer, we passed the extracted feature maps through $ 1 \times 1 $ convolutional layers to decrease the depth of the feature maps before concatenating with the decoder's outputs. In the end, the refinement block is responsible for capturing patches from the predicted alpha matte to refine them. }
\label{fig_method}
\end{figure*}

\textbf{Image matting} 
We can divide image matting literature into three main groups which are sampling-based methods \cite{chuang2001bayesian,feng2016cluster,gastal2010shared,he2011global,johnson2016sparse,karacan2015image,karacan2017alpha,ke2020guided}, propagation-based methods \cite{aksoy2017designing,aksoy2018semantic,bai2007geodesic,chen2013knn,grady2005random,levin2007closed,levin2008spectral,sun2004poisson}, and deep learning-based methods \cite{cho2016natural,shen2016deep,zhu2017fast,xu2017deep,chen2018semantic,lutz2018alphagan,zhang2019late,lu2019indices,li2020natural,Qiao_2020_CVPR,yu2020high,liu2020boosting,javidnia2020background,sengupta2020background,wu2020jmnet,forte2020f,ke2020green,liu2021towards, dai2021learning, liu2021tripartite, sun2021semantic, wei2021improved,zhong2021highly}. In deep learning-based methods, Convolutional Neural Networks (CNNs) and Generative Adversarial Networks (GANs) are proposed to perform alpha matte prediction for image matting \cite{cho2016natural,xu2017deep,lutz2018alphagan,lu2019indices,javidnia2020background,yu2020high,liu2021towards,zhong2021highly}. Besides, the attention mechanism increases the matting performance~\cite{li2020natural,Qiao_2020_CVPR}. Recently, trimap-free approaches become more and more important due to the difficulty of obtaining trimap \cite{zhang2019late,forte2020f,xu2021virtual}. 

\textbf{Portrait Matting} In \cite{shen2016deep}, a CNN-based end-to-end system is presented to produce an alpha matte for the portrait matting task. In \cite{chen2018semantic}, the key point of Semantic Human Matting (SHM) algorithm is to learn implicit semantic constraints from the data to use. 
Moreover, the authors provide a new dataset and a novel fusion strategy for the alpha matte. In \cite{wu2020jmnet}, end-to-end Joint Matting Network (JMNet) 
benefits from the pose of the human body to produce alpha matte and uses trimap refiner network to improve the sharpness. 
In \cite{liu2020boosting}, 
the proposed system contains three submodules that are predicting coarse semantic mask, improving the quality of the mask, and generating the final alpha matte. 
In \cite{sengupta2020background}, 
a trimap-free system takes an input image and the background of the same image without subject to generate alpha matte. 
To provide generalization, another matting network is trained in relation to the first network. In \cite{dai2020towards}, a light-weight method with two decoders and a single encoder is proposed. 
The task-specific decoders predict segmentation map and alpha matte using encoded semantic information. In \cite{ke2020green}, a matting objective decomposition network (MODNet) is proposed 
and a self-supervision-based strategy is applied to adjust it to the real-world scenario. 
In \cite{lin2020real}, 
the proposed method has two subnetworks and works in real-time with a high accuracy using HR images. While the first network takes an image and its background as input and generates four different outputs ---alpha matte, foreground residual, error map, and hidden---, the second network performs refinement. Besides, two large-scale datasets for video and image matting are presented. 
In \cite{xu2021virtual}, the proposed system works without additional input and the task is addressed as self-supervised multi-modality problem. The system utilizes 
depth-map, segmentation map, and interaction heat-map using three different encoders. A new dataset is also proposed in this work. In \cite{wang2021video}, the authors propose Consistency-Regularized Graph Neural Network to improve the temporal coherence during the video matting and they also collected a real-world dataset to evaluate the performance. In \cite{yu2021cascade}, Cascade Image Matting Network with Deformable Graph Refinement is presented to predict alpha matte automatically without using additional inputs. They predict the alpha matte from low resolution to high resolution. In \cite{mgm}, the proposed system takes a coarse mask to alleviate the alpha generation task. The system does not require a precise trimap but uses a general rough mask to guide the alpha matte prediction. In \cite{sun2021deep}, the authors propose a deep learning-based video matting system and present a novel spatio-temporal feature aggregation module. They also utilized frame-by-frame trimap annotations and contributed to the literature with a large-scale video matting dataset.

\section{Methodology}

We propose a two-stage approach to perform portrait matting task. 
Our model consists of two sub-models which are DeepLabv3+ \cite{deeplabv3_plus} for person segmentation and alpha matte generation network for alpha matte prediction. While the segmentation model takes a single RGB image to predict the segmentation map, the alpha matte generation network produces alpha matte using the input image as well as the predicted segmentation map. In the alpha generation network, there are two parallel similar encoder blocks, which are the content encoding block and the segmentation encoding block. While the content encoding block provides a feature representation for the input image, the segmentation encoding block encodes the depth-wise concatenation of the predicted segmentation map and foreground subject that is obtained by multiplying the input image and predicted segmentation map. Afterward, the outputs of both encoders are concatenated along with the depth. The concatenated feature representation passes through consecutive residual blocks and the decoder network to obtain the predicted alpha matte. Besides, there are skip connections between the decoder's layers and the encoders' layers. Since the concatenation of three different feature maps makes the feature representation too deep, we pass encoders' outputs through $ 1 \times 1 $ convolutions to reduce the dimension before concatenating with the decoder's layers' outputs. In the end, there is a small encoder-decoder network to enhance the predicted alpha matte by taking small patches from the borders of the subject since these regions are more likely to be inaccurate. The proposed system is shown in Figure~\ref{fig_method}. 

\textbf{Generators} While the content encoding block is responsible for encoding the input image to obtain a feature map, the segmentation encoding block provides a feature representation of the predicted segmentation map and foreground subject. The idea behind using separate encoders is to avoid vanishing the features of the segmentation map and the foreground subject. We also found that the segmentation map and extracted foreground subject provide complementary feature representations to the network. According to the experiments, we noticed that using a single encoder causes the vanishing of the feature representation of the additional inputs. Besides, we empirically realize that a less complex encoder is suitable to encode these additional inputs. Our generator is based on the U-Net generator \cite{unet} and it contains consecutive convolutional blocks to downsample the input image. After the encoding blocks, the features are concatenated and the final representation passes through the residual block. Later, the generator has a decoder module to produce an output alpha map by upsampling the residual output. Finally, $ 64 \times 64 $ size of consecutive patches in the border of a person's body are extracted from the predicted alpha matte to enhance the details by the refinement network, since the predictions tend to have more errors around the body. The generated image is expected to be the fine-grained alpha matte of the input image for the portrait matting problem. Besides, the skip connections encourage the network to keep information from both encoders.

\textbf{Multi-scale discriminator network} For the multi-scale discriminator \cite{pix2pixhd}, we provide an image pyramid using the original image and downsampled versions by a factor of two and a factor of four to obtain the same image on different scales. Therefore, this approach provides us to learn from a general perspective to finer details, since each discriminator has a different receptive field. Please note that all three discriminators are identical, though each discriminator works on a different scale. Since alpha matte does not contain a sufficient amount of useful representation, we decided to use a combination of the alpha matte and the extracted foreground subject, which is obtained by multiplying the alpha matte and the image, as an input to the discriminator network. For the real image, we extracted subjects using the images and the ground truth alpha matte, while we used the images and the predicted alpha matte to obtain fake images for the discriminator. Depth-wise concatenation of the three channels RGB image and one channel alpha matte is the input data of the discriminator.

\textbf{Loss functions} For the training of the alpha generation network, we used adversarial loss \cite{goodfellow2014generative}, perceptual loss \cite{johnson2016perceptual}, alpha loss, border loss, and alpha coefficient loss. In the perceptual loss \cite{johnson2016perceptual}, we utilized the VGG model \cite{simonyan2014very} to extract features. For this, we employed five different layers of the VGG model to obtain features for the generated image and the real image. We followed a similar pipeline as in \cite{johnson2016perceptual} to decide the layers to extract features. After that, we calculated a weighted sum of the L1 distances between features of the predicted alpha matte and the ground truth alpha matte for all extracted features. Besides, we applied the same loss for generated foreground subject and ground truth foreground subject that we obtained by multiplying the input image with predicted alpha matte and ground truth alpha matte, respectively. Then, we followed the same strategy to extract features and calculate the perceptual loss.

For the alpha loss, we followed a different strategy and calculated the L1 distance between the pixels that have only one or zero values in the pixel domain instead of calculating L1 distance between all pixels. The remaining pixels that have neither one nor zero values are considered by defining another loss based on L1 distance. Thus, we penalized the [0,1] pixels and the pixels between 0 and 1 separately, since they represent different cases and restrain each other when we consider them together. Please note that we also calculated both losses using alpha matte and foreground subjects as in the perceptual loss. Moreover, we proposed the border loss to penalize the area around the subject. For this, we generated border maps by applying morphological erosion and dilation operations separately. Then, we subtracted the eroded segmentation map from the dilated one. The final map represents the border area of the subject. During the training, we utilized this border map to calculate L1 loss for only the corresponding border pixels. The overall loss function is shown in Equation \ref{overall_loss} 

\begin{equation}
    \underset{G}{min}\underset{D_1, D_2, D_3}{max} \sum_{k=1,2,3} (L_{cGAN}(G,D_k) + \lambda L_{per}(G) + \atop \beta L_{alpha}(G)) + \gamma L_{border}(G) + \theta L_{ac}(G))
    \label{overall_loss}
\end{equation}
where $ L_{cGAN} $ represents conditional adversarial loss, $ L_{per} $ shows the perceptual loss, $ L_{alpha} $ indicates the alpha loss, $ L_{border} $ states the border loss, and $ L_{ac} $ expresses the alpha coefficient loss. Besides, $ \lambda, \beta, \gamma, \theta $ are coefficients that determine the effect of each losses over the total loss. According to our experiments on validation set, we empirically defined these values as 10, 25, 50, 25.

\subsection{Training procedure}

During the training, we did not train the segmentation network. Instead, we only trained the alpha generation network and the refinement network end-to-end. During the inference, the framework works end-to-end which means we provide an input image to the whole system and get an alpha matte for the corresponding input image. The input images are resized to $ 1280 \times 768 $ resolution before feeding the network. We used $ 10^{-4} $ learning rate for the generator and a ten times smaller learning rate ($ 10^{-5} $) for the discriminator to slow down the convergence of the discriminator since we empirically realized that discriminator converged too fast. We trained the alpha generation network with batch size of one as using one image in each batch causes better convergence \cite{forte2020f}. Besides, we utilized Adam optimizer~\cite{kingma2014adam} for the training of both models. We trained the discriminator one step for every five steps for the generator training.

\begin{table*}
\begin{center}
\begin{tabular}{|l|c|c|c|c|c|c|c|}
\hline
Method & Input & Dataset & MSE & MAE & SAD & Grad & Conn \\
\hline\hline
BGM-V2 \cite{lin2020real} & Image, background & AIM & 2.12 & 8.62 & 9.04 & 8.32 & 9.21 \\
\textbf{FBA} \cite{forte2020f} & Image, trimap & \textbf{AIM} & \textbf{0.40} & \textbf{3.79} & \textbf{3.98} & \textbf{1.19} & \textbf{3.11} \\
MODNet \cite{ke2020green} & Image & AIM & 21.65 & 32.36 & 33.93 & 44.24  & 35.45 \\
MGM~\cite{mgm} & Image & AIM & 1.48 & 5.96 & 6.21 & 4.74 & 6.55 \\
\textbf{Ours} & Image & \textbf{AIM} & \textbf{1.06} & \textbf{4.93} & \textbf{5.04} & \textbf{4.22} & \textbf{5.39} \\ 
\hline
\textbf{BMG-V2} \cite{lin2020real} & Image, background & \textbf{PM85} & \textbf{0.37} & \textbf{1.38} & \textbf{1.45} & \textbf{1.28} & \textbf{2.38}  \\
FBA \cite{forte2020f} & Image, trimap & PM85 & 1.01 & 2.43 & 2.55 & 3.50 & 2.75  \\
MODNet \cite{ke2020green} & Image & PM85 & 2.32 & 6.90 & 7.23 & 12.17 & 9.48   \\
MGM~\cite{mgm} & Image & PM85 & 0.38 & 2.77 & 2.91 & 1.32 & 2.04 \\
\textbf{Ours} & Image & \textbf{PM85} & \textbf{0.19} & \textbf{1.11} & \textbf{1.19} & \textbf{0.65} & \textbf{1.16} \\
\hline
BMG-V2 \cite{lin2020real} & Image, background & D646  & 0.98 & 4.60 & 4.83 & 3.78 & 5.30  \\
\textbf{FBA} \cite{forte2020f} & Image, trimap & \textbf{D646} & \textbf{0.44} & \textbf{3.10} & \textbf{3.25} & \textbf{1.70} &  \textbf{2.38} \\
MODNet \cite{ke2020green} & Image & D646 & 3.51 & 9.80 & 10.27 & 13.54 & 18.98  \\
MGM~\cite{mgm} & Image & D646 & 0.88 & 5.17 & 5.42 & 3.40 & 4.76 \\
\textbf{Ours} & Image & \textbf{D646} & \textbf{0.71} & \textbf{3.84} & \textbf{3.99} & \textbf{2.74} & \textbf{3.84}  \\
\hline
FBA \cite{forte2020f} & Image, trimap & PPM-100 & 0.96 & 2.24 & 2.41 & 4.20  & 2.70   \\
MODNet \cite{ke2020green} & Image & PPM-100 & 4.60 & 9.70 & 11.59 & 12.48 & 22.16  \\
MGM~\cite{mgm} & Image & PPM-100 & 1.15 & 5.07 & 5.31 & 5.04 & 5.29  \\
\textbf{Ours} & Image & \textbf{PPM-100} & \textbf{0.84} & \textbf{4.02} & \textbf{4.70} & \textbf{3.67} & \textbf{4.46}  \\
\hline
\end{tabular}
\end{center}
\caption{Quantitive evaluation on different datasets. Since PPM-100 dataset contains real-world images, we could not test BGM-V2 due to lack of background images. The corresponding MSE and MAE metrics are scaled by $ 10^3 $ to improve the readability. }
\label{table_results}
\end{table*} 


\section{Experimental Results}

\textbf{Datasets} In order to train our model, we used the combination of Adobe Image Matting (AIM) \cite{xu2017deep} and Distinctions (D646) \cite{Qiao_2020_CVPR} datasets to employ more data as well as increase the diversity. Since we focused on the portrait matting problem, we selected all images that contain persons for the training and test by following the same strategy in the portrait matting literature. In the end, there are 201 subjects in the AIM dataset and 363 subjects in the D646 dataset, making in total 564 subjects for the training set. We created the training set by following the standard strategy in the image matting literature for these datasets. For this, we combined each person in the training set with 100 different images of the MSCOCO dataset \cite{mscoco2014}. In the end, we have 56400 training images. For the test, we have four different test sets, namely, AIM \cite{xu2017deep}, PhotoMatte85 (PM85) \cite{lin2020real}, D646 \cite{Qiao_2020_CVPR}, and PPM-100 \cite{ke2020green}. We followed the same strategy and combined each person in the test set with 20 different background images of the PASCAL VOC dataset \cite{everingham2015pascal}. In the end, AIM contains 220 images (11 different subjects), PM85 includes 1700 images (85 different subjects), and D646 has 220 images (11 different subjects). The images in PPM-100 dataset have real backgrounds and there are 100 images in total. We evaluated our model on these four benchmark datasets and compared our results with the previous works. Please note that the training and test sets do not contain any common subjects, i.e. subject independent setup. The training and test subjects have already been listed for the corresponding datasets.

\textbf{Evaluation} We used mean squared error (MSE), mean absolute error (MAE), sum of absolute difference (SAM), gradient (Grad), and connectivity (Conn) metrics to evaluate our model as in the literature. 
For comparison, we chose publicly available SOTA methods, namely MODNet~\cite{ke2020green}, BGM-V2~\cite{lin2020real}, FBA~\cite{forte2020f}, MGM~\cite{mgm} and we tested them on the test sets in order to perform a fair comparison since different backgrounds may change the models' performances. Please note that we calculated these metrics over the whole image, and MSE and MAE scores are scaled by $10^3$ to improve the readability. Besides, we performed a user study to compare our results with the other studies. To perform this study, we combined the extracted subjects with a green background and showed these images to the participants.


\subsection{Results}

In this section, we present the experimental results and compare them with the recent SOTA works in the background matting literature, MODNet \cite{ke2020green}, FBA \cite{forte2020f}, BGM-V2 \cite{lin2020real}, and MGM~\cite{mgm}. Please note that, while our method and MODNet take an input image to generate alpha matte for portrait matting, BGM-V2 requires the original background image without subject as an additional input and FBA expects trimap to identify background, foreground, and unknown areas in addition to the original input image. Besides, MGM~\cite{mgm} requires a segmentation mask as our alpha matte generation network.

\begin{figure*}
\begin{center}
\begin{tabular}{ccccccc}
\includegraphics[width=2.0cm]{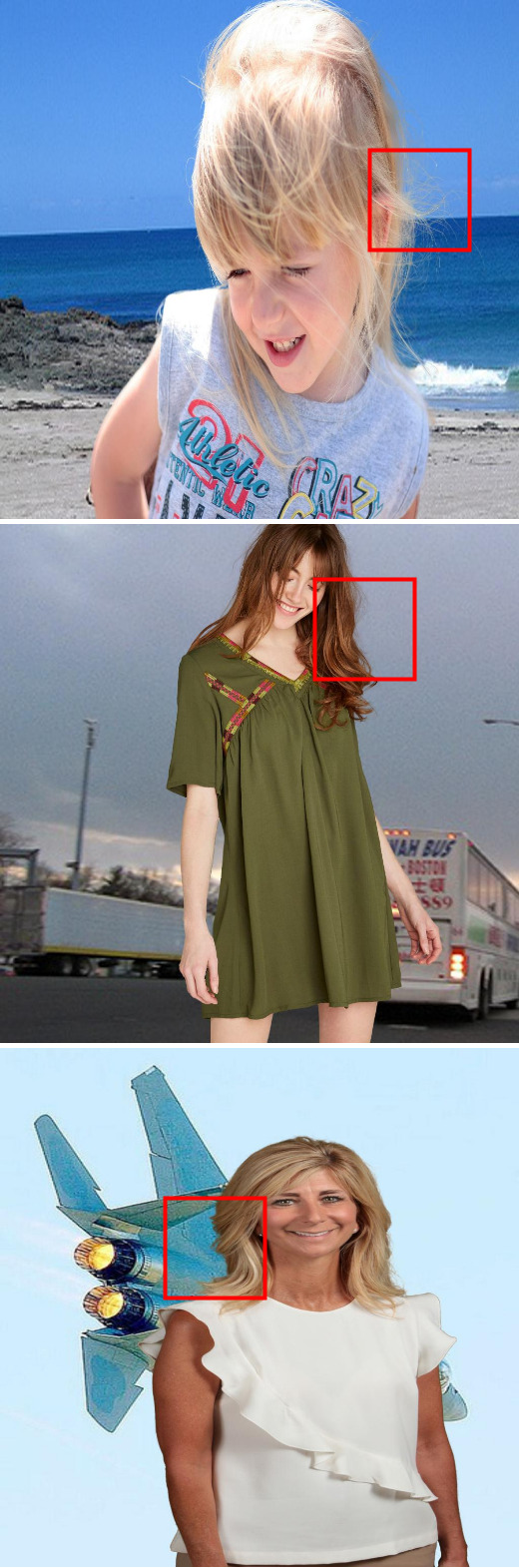}&
\includegraphics[width=2.0cm]{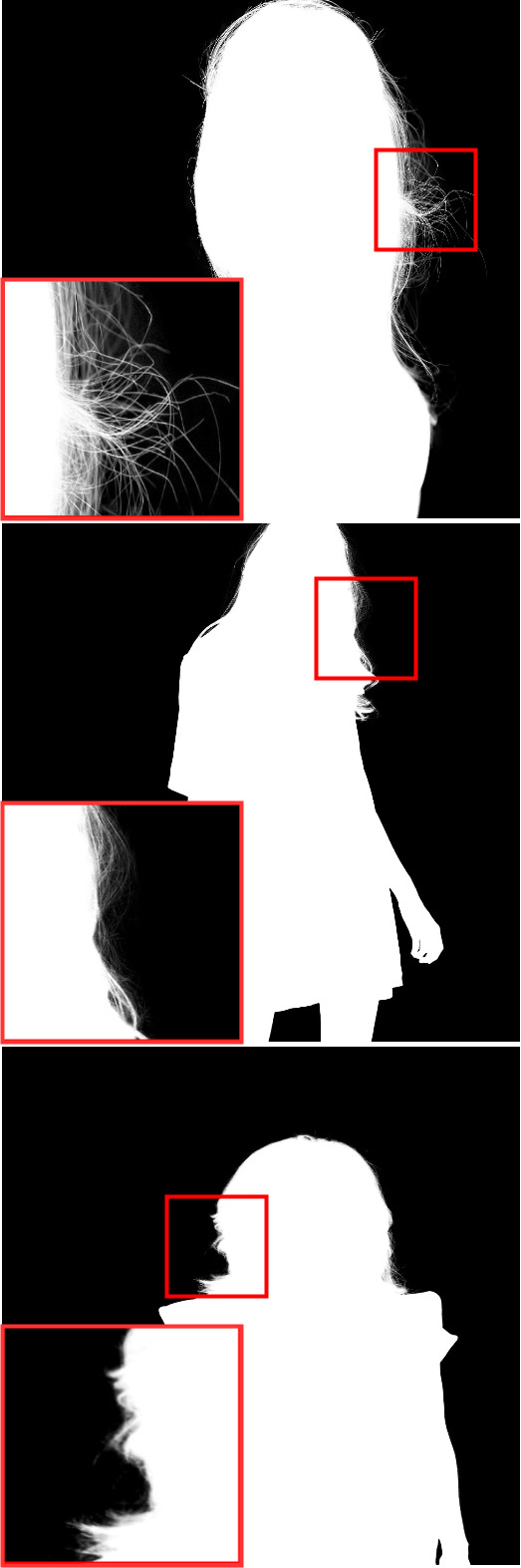}&
\includegraphics[width=2.0cm]{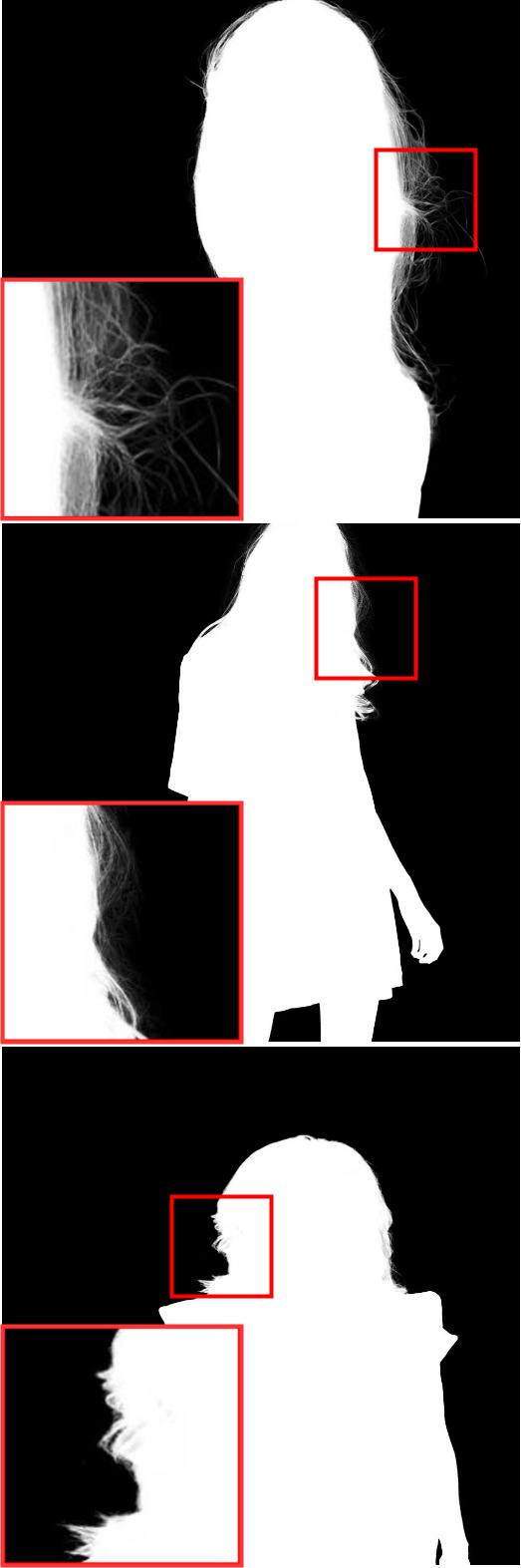}&
\includegraphics[width=2.0cm]{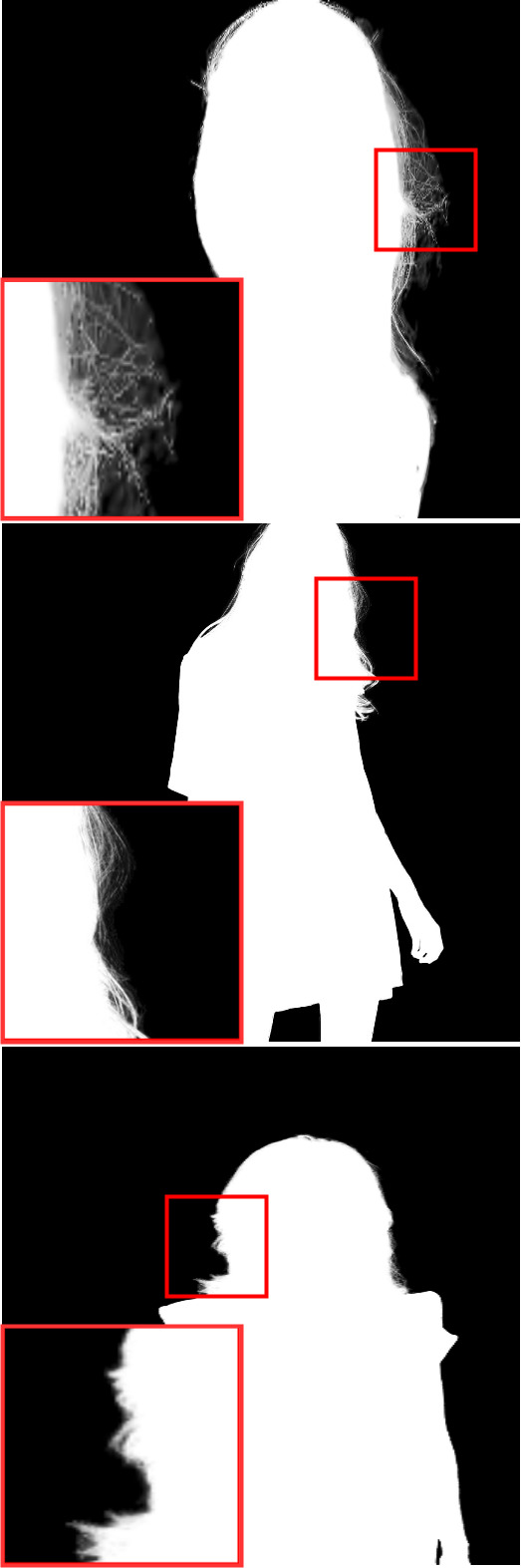}&
\includegraphics[width=2.0cm]{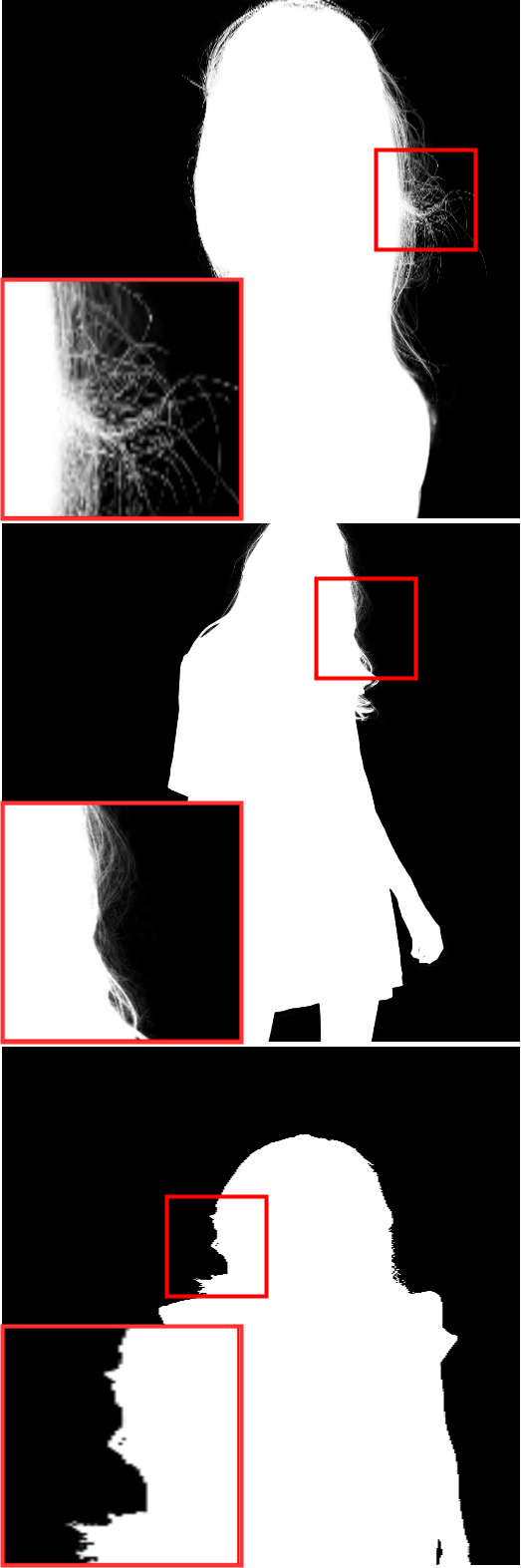}&
\includegraphics[width=2.0cm]{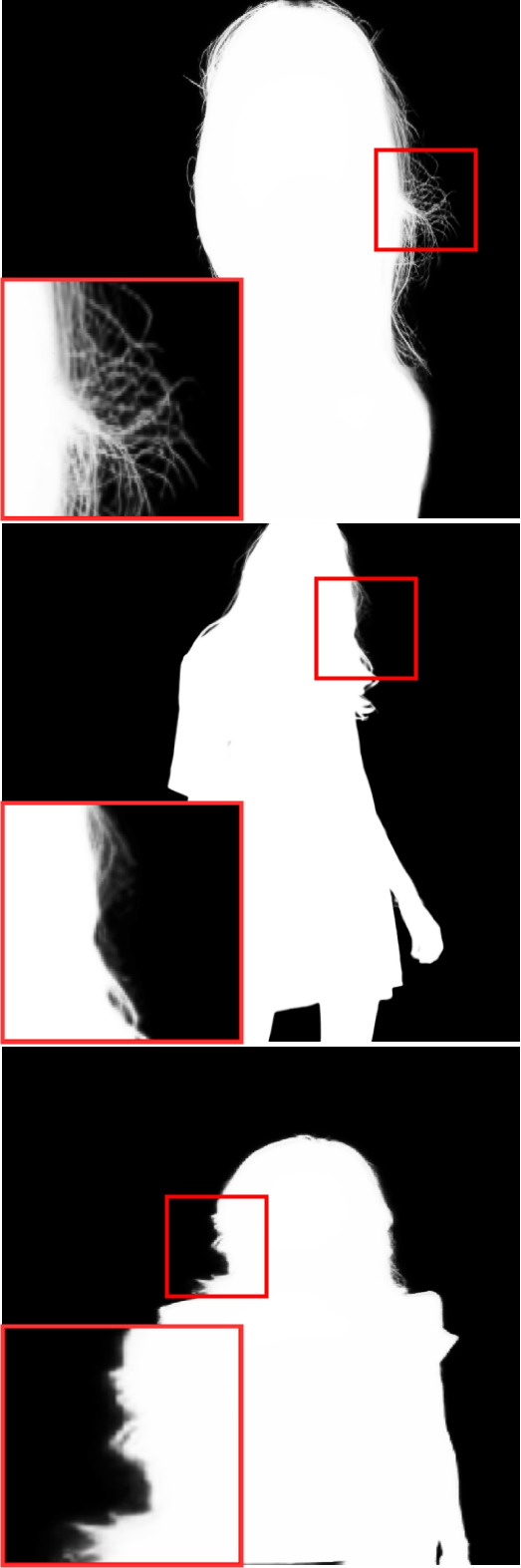}&
\includegraphics[width=2.0cm]{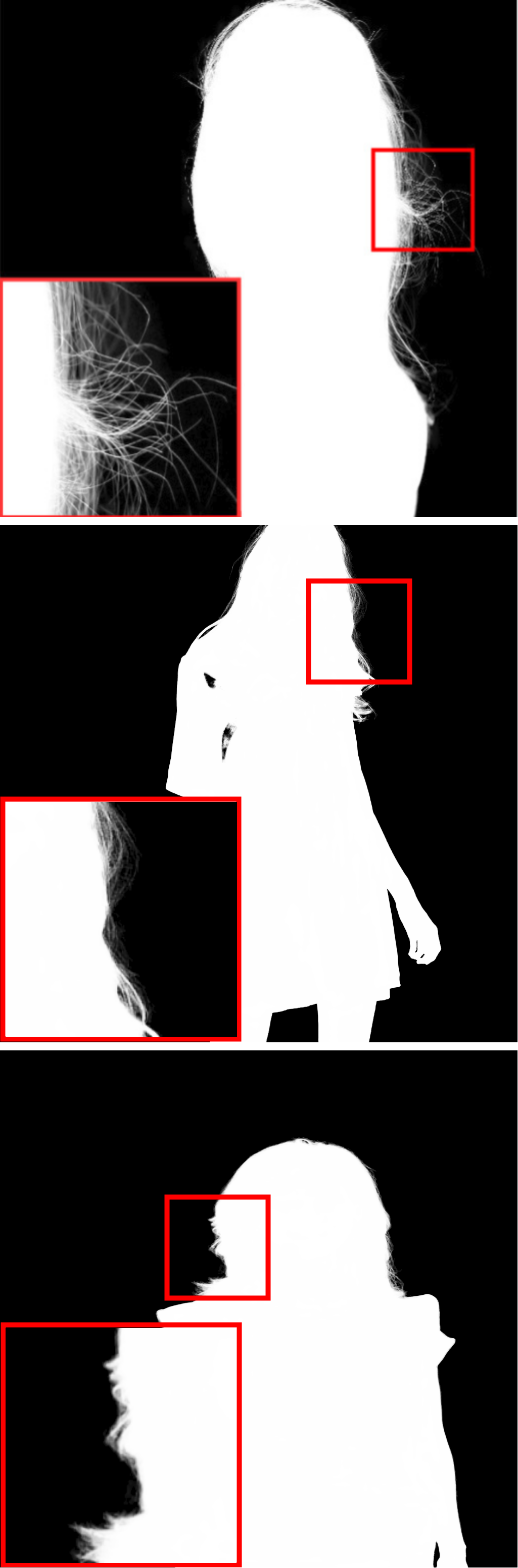}\\
Input & GT & Ours & BGM-V2 & FBA & MODNet & MGM
\end{tabular}
\end{center}
\caption{Qualitative comparison. Rows represent AIM, D646, PM85 datasets, respectively.}
\label{fig_results}
\end{figure*}

\textbf{Quantitative evaluation} Experimental results are shown in Table \ref{table_results}. We evaluated all models under the same conditions, e.g., using background image and resolution. According to the experimental results presented in Table \ref{table_results}, our model surpassed the performance of MODNet and MGM, which do not use any additional inputs, on four different benchmarks. Other methods in the table ---BGM-V2 and FBA--- benefit from additional input such as the background of the input image and a trimap. These additional inputs make the task easier and more accurate results are likely to be obtained, since the background image is the same one as the original input image, and trimap identifies most of the area on the image as foreground and background. On the AIM dataset, our model is found superior to the BGM-V2 in all metrics. However, the FBA method achieves the best performance on this test set. On the PM85 dataset, our proposed model outperforms all methods and gets the SOTA result. In the D646 benchmark, we again outperform the MODNet, MGM, and BGM-V2. The FBA reaches the best performance. However, it is slightly better than our method and our results are quite acceptable when compare with the FBA. Please note that since each study creates the test setup with a different set of background images, the presented scores may show differences.

As previously stated, while our approach does not take any input in addition to the original image, the FBA method takes trimap and the BGM-V2 method takes the background of the original input image that does not contain the subject itself. However, they are too sensitive to these additional inputs. For instance, if there are any dissimilarities in the background image such as translation, BGM-V2 cannot produce a proper output and generates a completely corrupted prediction instead. Similarly, FBA is sensitive to the trimap input. In addition to all these cases, our model and all other models are sensitive to the background of the input image according to the findings of our detailed experiments. It indicates that the alpha matte prediction performance of the models for the same subject can considerably change according to the background of the input image. The illumination conditions, the color distribution, and the existence of multiple subjects on the image affect the alpha matte prediction performance. For the PPM-100 dataset, since the images are real-world images, there are no background images without the subject. Therefore, we could not test the BGM-V2 model on this dataset.

\begin{figure*}
\begin{center}
\begin{tabular}{cccccc}
\includegraphics[width=2.4cm]{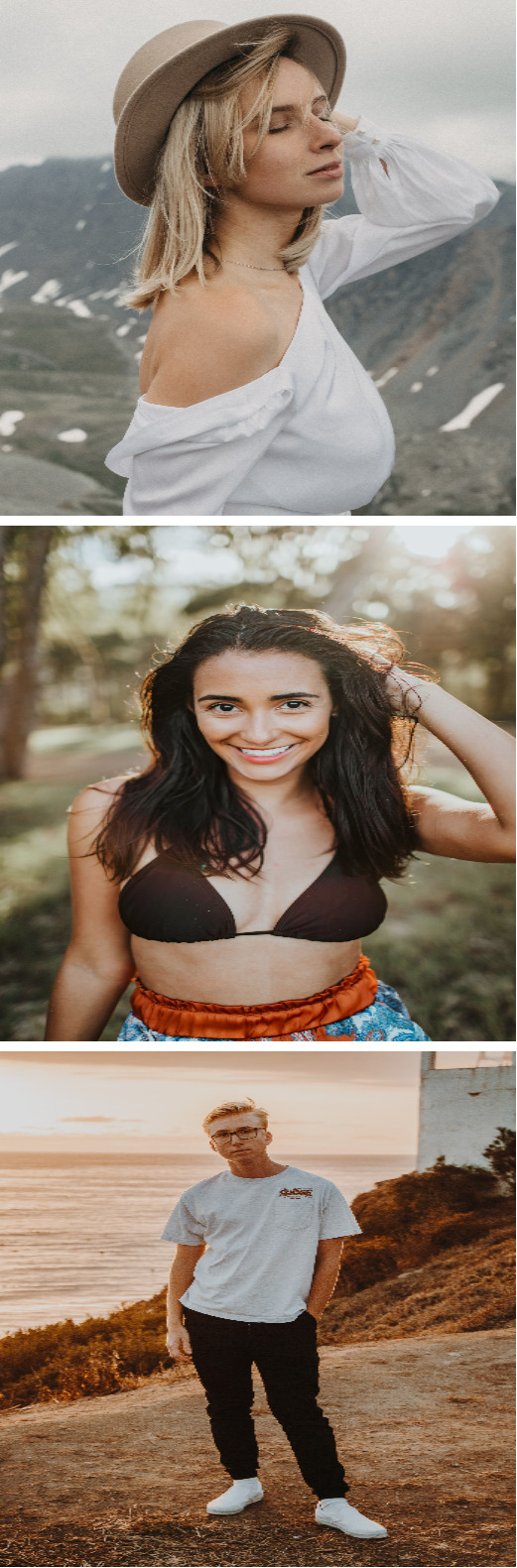}&
\includegraphics[width=2.4cm]{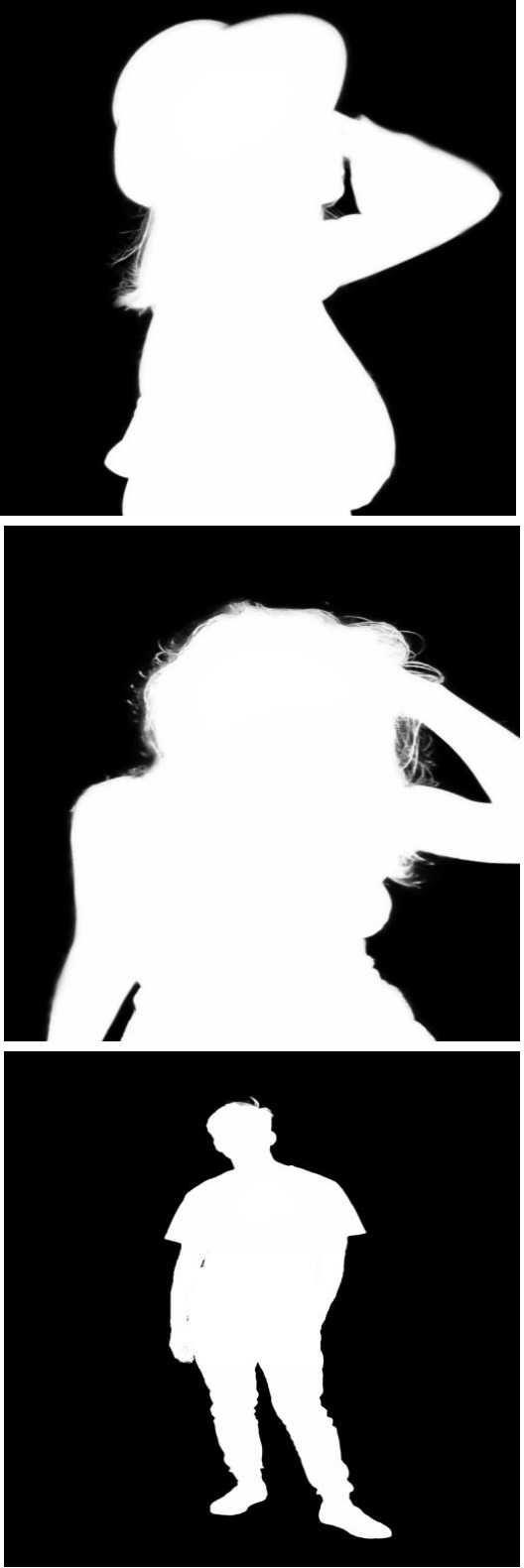}&
\includegraphics[width=2.4cm]{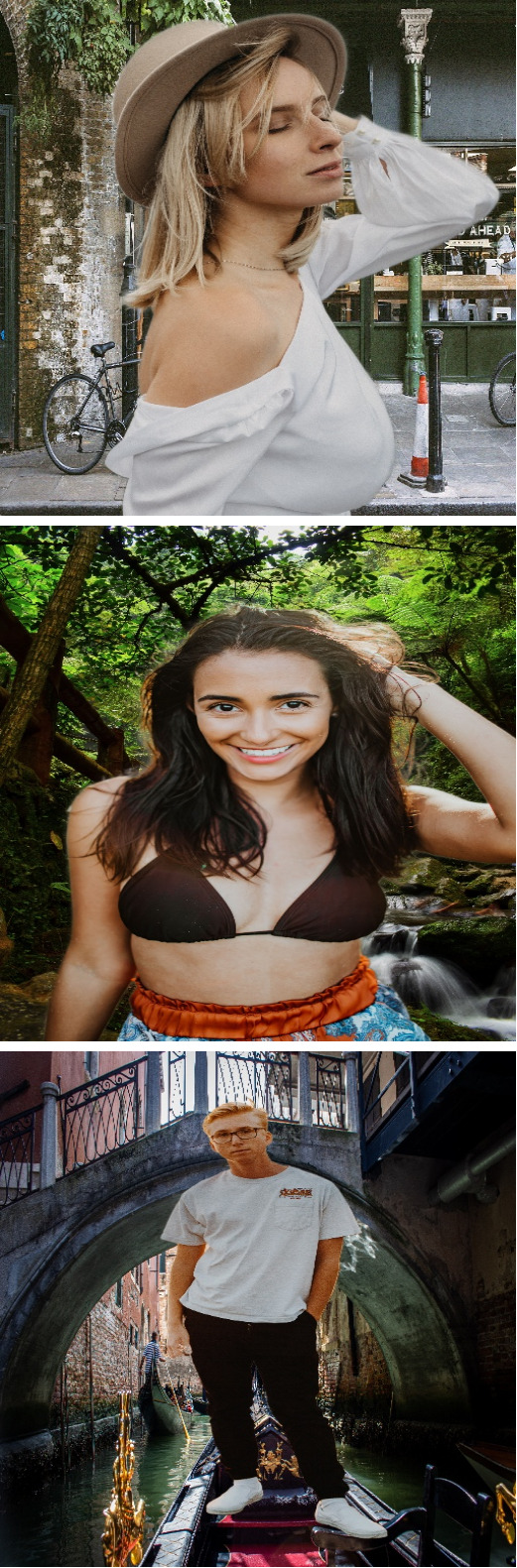}&
\includegraphics[width=2.4cm]{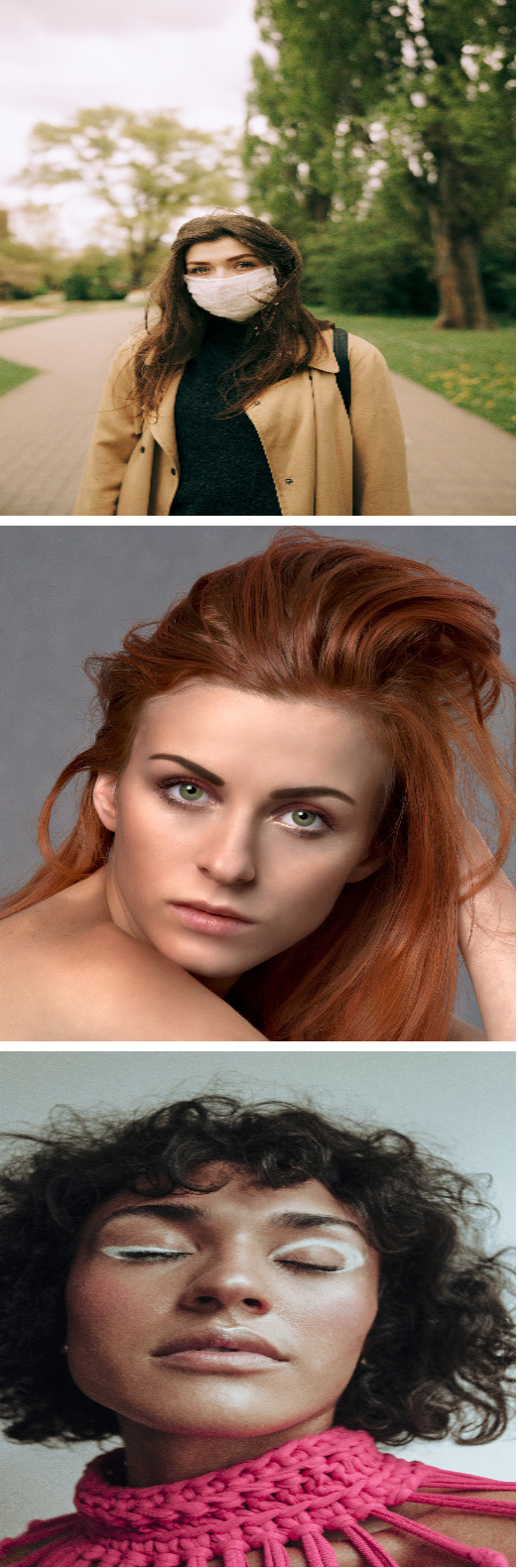}&
\includegraphics[width=2.4cm]{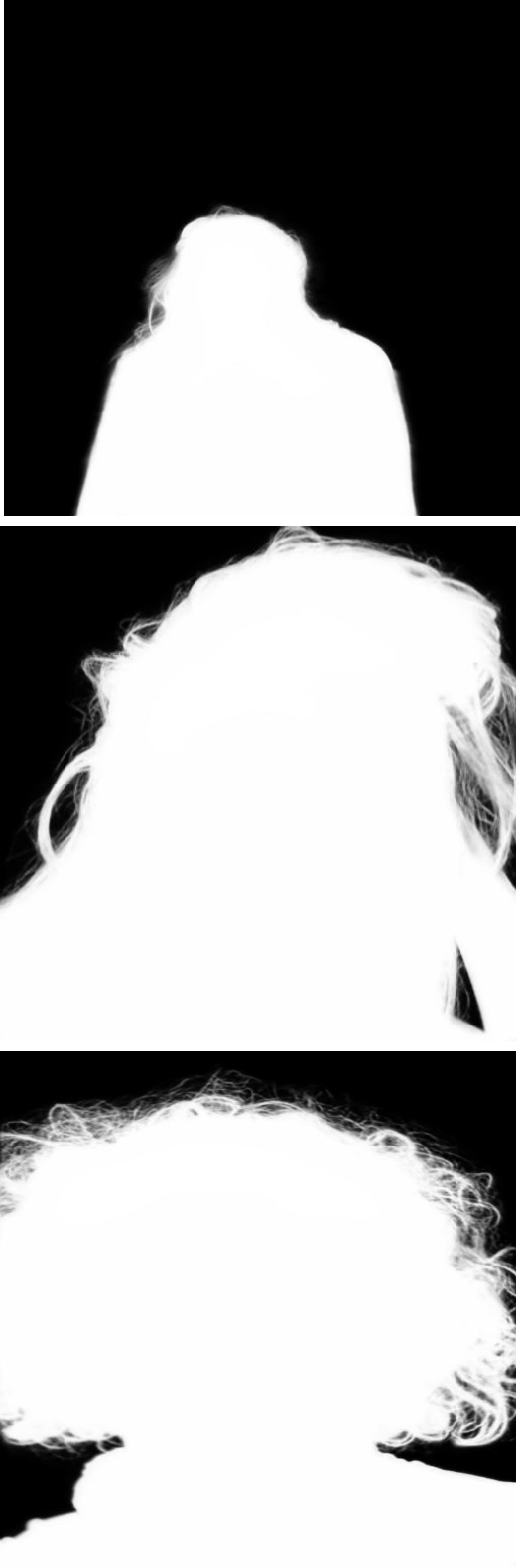}&
\includegraphics[width=2.4cm]{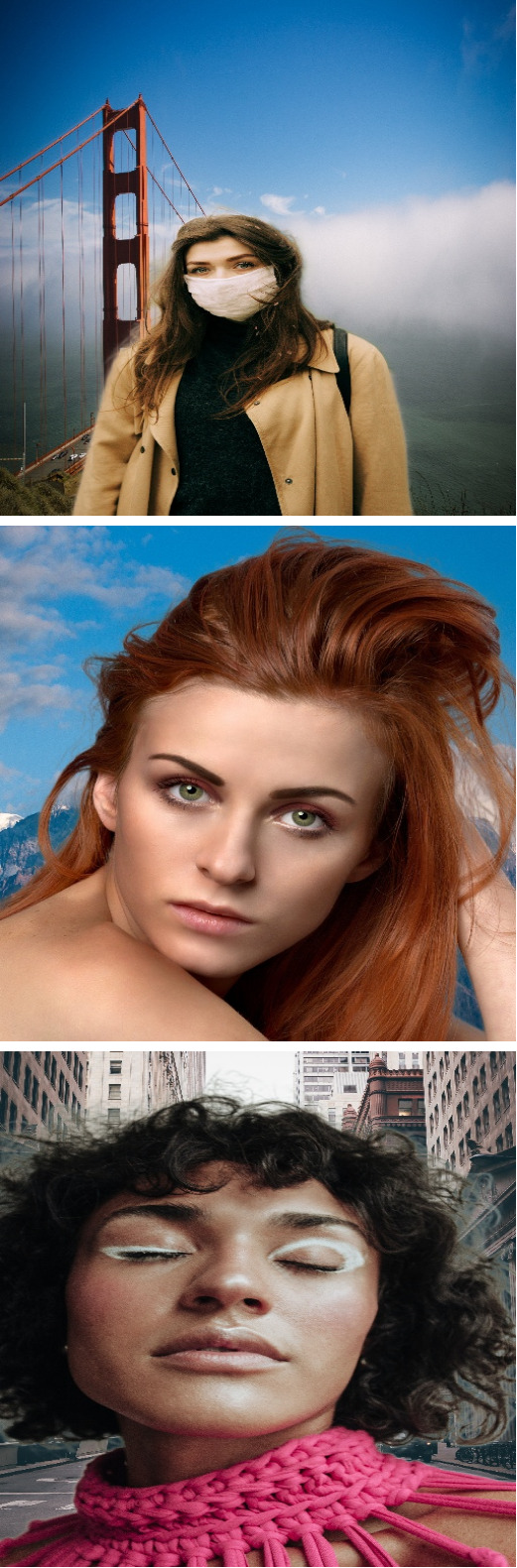}\\
Input & Alpha & Combined & Input & Alpha & Combined
\end{tabular}
\end{center}
\caption{We tested our models on the real images that were collected from the web. In the end, we changed the backgrounds with arbitrary backgrounds using the predicted alpha matte to show the application of the system.}
\label{fig_results_application}
\end{figure*}

\textbf{Qualitative evaluation} In Figure \ref{fig_results}, we present our results, input image, ground truth, and the outputs of the other models for three benchmark datasets; AIM, D646, and PM85. We generated outputs with our model, MGM, and MODNet without additional inputs. However, BGM-V2 method needs the same background of the input image and FBA requires trimap for the corresponding input data. For BGM-V2, we provided the background image that we used during the preparation of the test data. Since D646, PM85, and PPM-100 datasets do not include trimaps, we created different trimaps by using erosion and dilation operations to evaluate FBA and present the best scores. According to the figure, our results are almost the same as the ground truth data, especially for the challenging part, such as hair. Besides, although all models perform quite well, the differences between them are in the details, particularly around the borders of the subjects. Moreover, we randomly collected images from the web and we run our model over them to present the performance of the system on the real-world images. The corresponding outputs are presented in Figure \ref{fig_results_application}. The Alpha column contains the predicted alpha matte and the combined column includes the combination of an arbitrary background image and the extracted subject by using the predicted alpha matte.


We also performed a user study and asked 30 different participants to compare all results according to the quality of the images to measure the matting performance. We used randomly selected sample images from all four benchmark test sets. We present the results in Table \ref{table_user_study}. We have five different levels of score which are \textit{much better, better, same, worse, } and \textit{much worse} to compare our results with four different methods. The scores indicate how much the output image of our method is better or worse than the output of other methods. For the comparison, we extracted subjects from the image using predicted alpha matte and combined with a green background to make the details of the subject more visible for the users. During the survey, we showed the original input image and the combination of a green background and outputs of the models. 
We utilized 8 subjects for each test benchmark, except PPM100 since we could not test the BGM-V2 model on them, for the user study 
and we made pairs with our results and other results to show them to the participants. In total, we have 24 images for each model to create questions. According to the table, our model overperforms the MODNet and it is slightly better than BGM-V2 and MGM. On the other hand, participants could not easily distinguish our results and FBA results and majority, 52.44\%, said they are the same. 

\begin{table}
\begin{center}
\begin{tabular}{@{}l|c|c|c|c@{}}
\hline
Score            & MODNet  & BGM-V2 & FBA  & MGM  \\
\hline\hline
Much better & 41.55\% & 10.45\% & 4.23\% & 8.25\% \\
Better      & 29.22\% & 32.67\% & 16.61\% & 30.27\% \\
Same        & 19.15\% & 39.86\% & 52.44\% & 41.02\% \\
Worse       & 8.11\%  & 16.33\% & 22.47\% & 18.20\% \\
Much worse  & 1.94\%  & 0.69\%  & 3.58 \% & 2.26\% \\
\hline
\end{tabular}
\end{center}
\caption{User study using all three benchmarks. We compared our model with MODNet \cite{ke2020green}, BGM-V2 \cite{lin2020real}, FBA \cite{forte2020f}, and MGM~\cite{mgm}. The scores demonstrate how much our result is better or worse than the other results.}
\label{table_user_study}
\end{table}

\subsection{Ablation study}


\textbf{Loss functions} We performed an ablation study to evaluate the effects of different parameters on the performance. We first investigated the loss functions and then utilized data type for the losses. In the first part of Table \ref{table_ablation_study_loss}, we show used loss functions for the training as well as corresponding MSE values on the AIM test set. 
It is observed that each employed additional loss contributes significantly to the prediction performance of the model. In the bottom part of Table \ref{table_ablation_study_loss}, we present the effect of using the alpha matte and the foreground subject in the loss functions. $ \alpha $ means that we only utilized predicted alpha matte and ground truth alpha matte. $ \alpha $ and $ F $ represent that we extracted the subject from the input image with predicted alpha matte and ground truth alpha matte to obtain predicted and real foreground subjects. Then, we employed these outputs to calculate loss functions for the corresponding case. While using alpha matte helps to penalize the difference between predicted and ground truth alpha matte, using foreground subject provides more information to the network, since it contains much more details and semantic information than the alpha matte. According to the results, MSE scores indicate that using the foreground subject in addition to the alpha matte enables the network to produce a more accurate map.

\begin{table}
\begin{center}
\begin{tabular}{l|c}
\hline
Loss        & MSE  \\
\hline\hline
$ L_{cGAN} + L_{alpha} $ & 7.24 \\
$ L_{cGAN} + L_{per} + L_{alpha} $ & 3.78 \\
$ L_{cGAN} + L_{per} + L_{alpha} + L_{border} $ & 1.76 \\
$ L_{cGAN} + L_{per} + L_{alpha} + L_{border} + L_{ac} $ & 1.06 \\
\hline
$ \alpha $ & 3.14 \\
$ \alpha, F $ & 1.06 \\
\hline
\end{tabular}
\end{center}
\caption{Ablation study for the loss functions. We repeated the training of the alpha matte generation network using a combination of different loss functions and we present MSE results on AIM test set in the top part of the table. We additionally show the results with all loss functions by using only alpha matte and using foreground subject and alpha matte together in the loss functions. }
\label{table_ablation_study_loss}
\end{table}




\textbf{Modules} We further examined the effect of the segmentation encoder block and the refinement network. According to the results in Table \ref{table_ablation_network}, both the segmentation encoding block and the refinement network are significantly useful to improve the performance of the proposed method. Because, the segmentation encoding block improves the representation of the segmentation area by providing the encoded feature representation to the residual block, while the refinement network enhances the alpha matte prediction performance by focusing on the challenging parts. 

\textbf{Input type} We analyzed how using foreground subject in the generator and discriminator as input affects the performance. 
The results in Table \ref{table_ablation_input_type_and_D} indicate that providing a foreground subject in addition to the segmentation map, which we obtained by multiplying the input and the predicted segmentation map, increases the performance since it provides a more effective feature representation. Similarly, concatenation of the alpha matte and extracted foreground subject provides a more useful representation to the discriminator that yields improvement in the performance. Please note that we evaluated the proposed system on the test set of the AIM dataset.

\begin{table}
\begin{center}
\begin{tabular}{l|c}
\hline
Cases        & MSE \\
\hline\hline
Base model & 2.20 \\ 
Base model + SE block & 1.57 \\ 
Base model + SE block + refinement network & 1.06 \\
\hline
\end{tabular}
\end{center}
\caption{Ablation study for the architecture. We individually investigated the effect of the segmentation encoding block and the refinement module. The experiments are performed on the AIM dataset.}
\label{table_ablation_network}
\end{table}

\textbf{Limitations} Our work is sensitive to the performance of the segmentation network. A poor quality segmentation output causes a less accurate outcome at the end of the alpha matte network due to a lack of visual representation of the subject. Besides, due to consecutive residual blocks, the model is not able to run in real-time.

\section{Conclusion}

\begin{table}
\begin{center}
\begin{tabular}{l|c}
\hline
Cases        & MSE \\
\hline\hline
Segmentation map & 1.86 \\
Segmentation + Foreground & 1.41 \\
\hline
Alpha matte + Foreground & 1.06 \\
\hline
\end{tabular}
\end{center}
\caption{Ablation study for the input type of the generator and discriminator. While the first part shows the input of the segmentation encoding block in the generator, the second part of the table indicates the input type of the discriminator network. }
\label{table_ablation_input_type_and_D}
\end{table}

In this work, we proposed a conditional GAN-based additional input-free approach to perform the portrait matting task. We addressed the problem as two different sub-problems. In the first step, we proposed to use DeepLabV3+ person segmentation model to generate a coarse segmentation map from an arbitrary input image. In the second step, this output and the original image are sent to the alpha generation network to generate the alpha matte. We presented the segmentation encoding block that encodes the combination of the predicted segmentation map and the foreground object. In the end, we have a refinement network to enhance the prediction quality by capturing several patches from the predicted alpha matte in the border area of the subject. Besides, we proposed border loss to penalize challenging parts around the subject and we also presented alpha coefficient loss to measure only the pixels in the alpha matte that the alpha coefficients are neither zero nor one. To handle the domain shift problem, we combined two important training datasets to increase the amount of data as well as the diversity. Experimental results indicate that using border loss and alpha coefficient loss improved the accuracy of the model and combining two datasets increased the generalization capacity. It is also observed that encoding the combination of the segmentation map and the foreground subject by the segmentation encoding block provided more useful features than encoding only the segmentation map. We also found out that the same outcome is also correct for the discriminator. When we provided the prediction output and the foreground subject, the discriminator worked better and was more stable. In future work, it is necessary to focus on the performance to make our model be able to run in real-time with sequential data in order to increase the possibility of usage in the real world. A possible scenario is to utilize the system by eliminating the background to provide privacy in the human-robot interaction.

\textbf{Acknowledgement.} The project on which this report is based was funded by the Federal Ministry of Education and Research~(BMBF) of Germany under the number~01IS18040A.

{\small
\bibliographystyle{ieee_fullname}
\bibliography{PaperForReview}
}

\end{document}